\pdfoutput=1
\documentclass[11pt]{article}

\usepackage[final]{emnlp2021}

\usepackage{times}
\usepackage{latexsym}
\usepackage{hyperref}

\usepackage[T1]{fontenc}
\usepackage[utf8]{inputenc}

\usepackage{microtype}
\usepackage{url}
\usepackage{makecell}
\usepackage{multirow}
\usepackage{graphicx}
\usepackage{caption}
\usepackage{subcaption}
\usepackage{amsmath}
\usepackage{bbm}
\usepackage{booktabs}
\usepackage{algorithmic}
\usepackage{amssymb}
\usepackage{bm}
\usepackage[ruled,vlined,linesnumbered]{algorithm2e}
\usepackage{enumitem}
\usepackage{pifont}

\newcommand{\xmark}{\ding{55}}%
\newcommand{\dataset}{Visual News}
\newcommand{\goodnews}{GoodNews}
\newcommand{\NYT}{NYTimes800k}

\title{\dataset: Benchmark and Challenges in News Image Captioning}

\author{
        Fuxiao Liu \\ University of Maryland \\ fl3es@umd.edu \And
        
        Yinghan Wang\Thanks{\space Work completed before joining Amazon.} \\ Amazon Alexa \\ yinghanw@amazon.com \AND 
         
        Tianlu Wang \\ University of Virginia \\ tianlu@virginia.edu \And
         
        Vicente Ordonez \\ Rice University \\ vicenteor@rice.edu}

\begin{document}
\maketitle
\begin{abstract}
We propose \dataset~Captioner, an entity-aware model for the task of news image captioning. We also introduce \dataset, a large-scale benchmark consisting of more than one million news images along with associated news articles, image captions, author information, and other metadata.
Unlike the standard image captioning task, news images depict situations where people, locations, and events are of paramount importance. 
Our proposed method can effectively combine visual and textual features to generate captions with richer information such as events and entities. More specifically, built upon the Transformer architecture, our model is further equipped with novel multi-modal feature fusion techniques and attention mechanisms, which are designed to generate named entities more accurately. 
Our method utilizes much fewer parameters while achieving slightly better prediction results than competing methods.
Our larger and more diverse \dataset~dataset further highlights the remaining challenges in captioning news images. 
\end{abstract}

\section{Introduction}
Image captioning is a language and vision task that has received considerable attention and where important progress has been made in recent years~\cite{vinyals2015show,fang2015captions,xu2015show,lu2018neural,anderson2018bottom}. 
This field has been fueled by recent advances in both visual representation learning and text generation, and also by the availability of image-text parallel corpora such as the Common Objects in Context (COCO) Captions dataset~\cite{chen2015microsoft}.

\begin{figure}[t]
    \centering
      \includegraphics[width=0.46\textwidth]{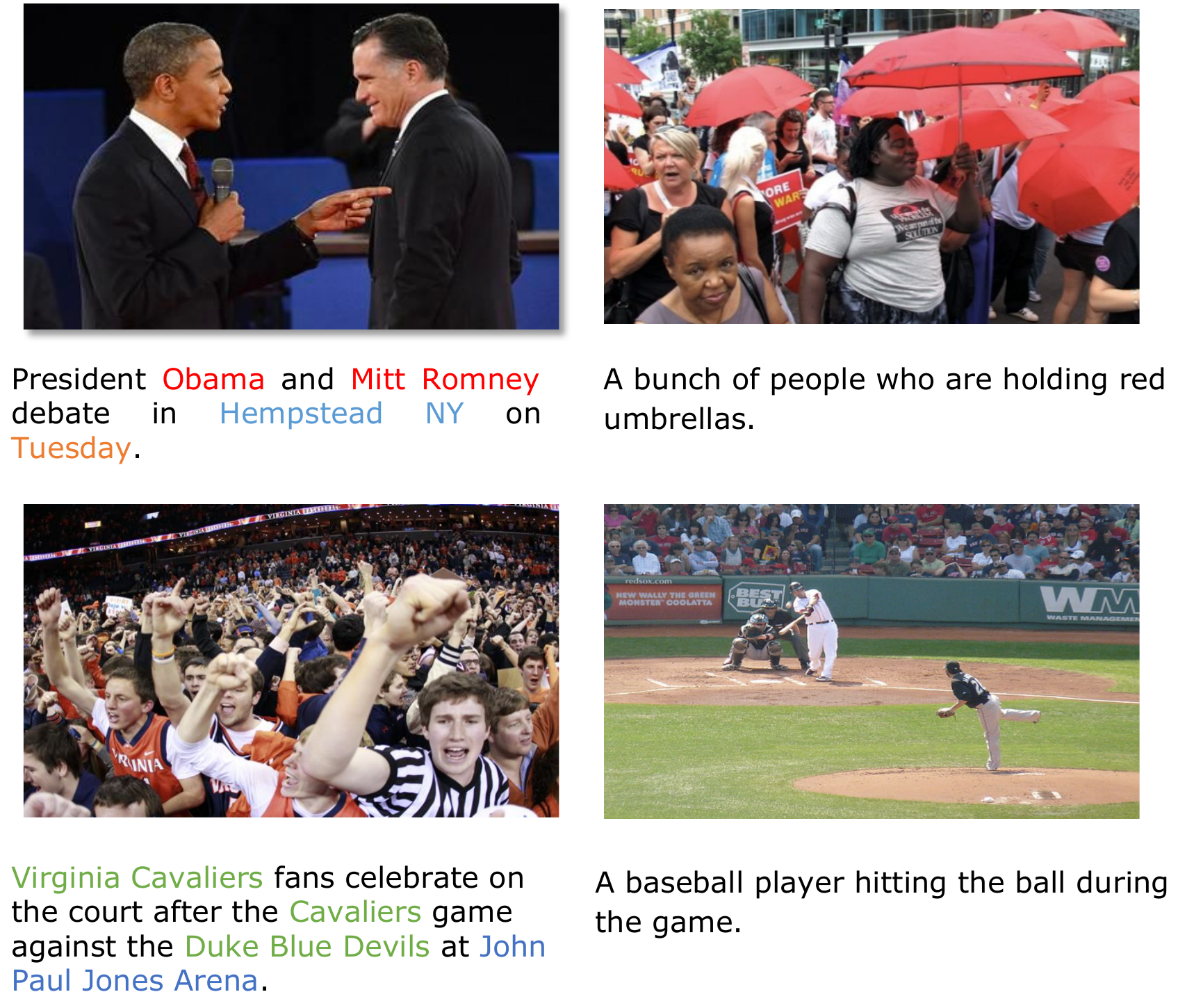}
     %\vspace{-3pt}
    \caption{Examples from our \dataset~dataset (left) and COCO~\cite{chen2015microsoft} (right). \dataset~provides more informative captions with name entities, whereas COCO contains more generic captions.
    }
    \vspace{-0.2in}
    \label{fig:lead}
\end{figure}

While COCO contains enough images to train reasonably good captioning models, it was collected so that objects depicted in the images are biased toward a limited set of everyday objects. Moreover, while it provides high-quality human annotated captions, these captions were written so that they are descriptive rather than interpretative, and referents to objects are generic rather than specific. For example, a caption such as~\textit{``A bunch of people who are holding red umbrellas.''} properly describes the image at some level to the right in Figure~\ref{fig:lead}, but it fails to capture the higher level situation that is taking place in this picture~i.e.~\textit{``why are people gathering with red umbrellas and what role do they play?''} This type of language is typical in describing events in news text. While a lot of work has been done on news text corpora such as the influential Wall Street Journal Corpus~\cite{paul1992design}, there have been considerably fewer resources of such news text in the language and vision domain.

In this paper, we introduce \dataset, a dataset and benchmark containing more than one million publicly available news images paired with both captions and news article text collected from a diverse set of topics and news sources in English (The Guardian, BBC, USA TODAY, and The Washington Post). 
By leveraging this dataset, we focus on the task of News Image Captioning, which aims at generating captions from both input images and corresponding news articles. We further propose \dataset~Captioner, a model that generates captions by attending to both individual word tokens and named entities in an input news article text, and localized visual features.   

News image captions are typically more complex than pure image captions and thus make them harder to generate.
News captions describe the contents of images at a higher degree of specificity and as such contain many named entities referring to specific people, places, and organizations. Such named entities convey key information regarding the events presented in the images, and conversely events are often used to predict what types of entities are involved. e.g.~if the news article mentions a baseball game then a picture might involve a baseball player or a coach, conversely if the image contains someone wearing baseball gear, it might imply that a game of baseball is taking place. As such, our \dataset~Captioner model jointly uses spatial-level visual feature attention and word-level textual feature attention.

More specifically, we adapt the existing Tranformer~\cite{vaswani2017attention} to news image datasets by integrating several critical components. To effectively attend to important named entities in news articles, we apply the Attention on Attention technique on attention layers and introduce a new position encoding method to model the relative position relationships of words. We also propose a novel Visual Selective Layer to learn joint multi-modal embeddings. To avoid missing rare named entities, we build our decoder upon the pointer-generator model.
% Another challenge in news image captioning is that named entities contain multiple words which are , e.g.~``Hillary Clinton'' or ``New York'' a new position embedding mechanism to. 
News captions also contain a significant amount of words falling either in the long tail of the distribution or resulting in out-of-vocabulary words at test time. 
% A lot of such tokens are references to entities such as the name of people, locations, or organizations with limited media exposure.
In order to alleviate this, we introduce a tag cleaning post-processing step to further improve our model.

% To better generate these multi-token entities, we propose a novel Entity-Aware module which combines information from the state vector corresponding to the previous token to generate the current token. To better capture information from named entities, we add an Entity-Guide attention layer which specifically injects named entity features. 

Previous works \cite{lu2018entity, biten2019good} have attempted news image captioning by adopting a two-stage pipeline. They first replace all specific named entities with entity type tags to create templates and train a model to generate template captions with fillable placeholders. Then, these methods search in the input news articles for entities to fill placeholders. Such approach reduces the vocabulary size and eases the burden on the template generator network. However, our extensive experiments suggest that template-based approaches might also prevent these models from leveraging contextual clues from the named entities themselves in their first stage.

Our main contributions can be summarized as:

\begin{itemize}[nosep,leftmargin=*]
    \item We introduce \dataset, the largest and most diverse news image captioning dataset and study to date, consisting of more than one million images with news articles, image captions, author information, and other metadata.
    %\vspace{0.05in}
    \item We propose \dataset~Captioner,~a captioning method for news images, showing superior results on the GoodNews~\cite{biten2019good}, NYTimes800k~\cite{tran2020transform} and \dataset~datasets with much fewer parameters than competing methods.
    \item We benchmarked both template-based and end-to-end captioning methods on two large-scale news image datasets, revealing the challenges in the task of news image captioning.
    % \item Our proposed Entity-Aware module and Entity-Guide attention layer that improves the generation of named entities for new image captions. 
    %\vspace{0.05in}
    % \item We demonstrate that training a model on raw text is more beneficial than aiming to generate templates and then filling named entities in a second stage.
\end{itemize}

Visual News text corpora, public links to download images, and further code and data are publicly available.~\footnote{\url{https://github.com/FuxiaoLiu/VisualNews-Repository}
}

\begin{figure}[t]
    \centering
      \includegraphics[width=0.47\textwidth]{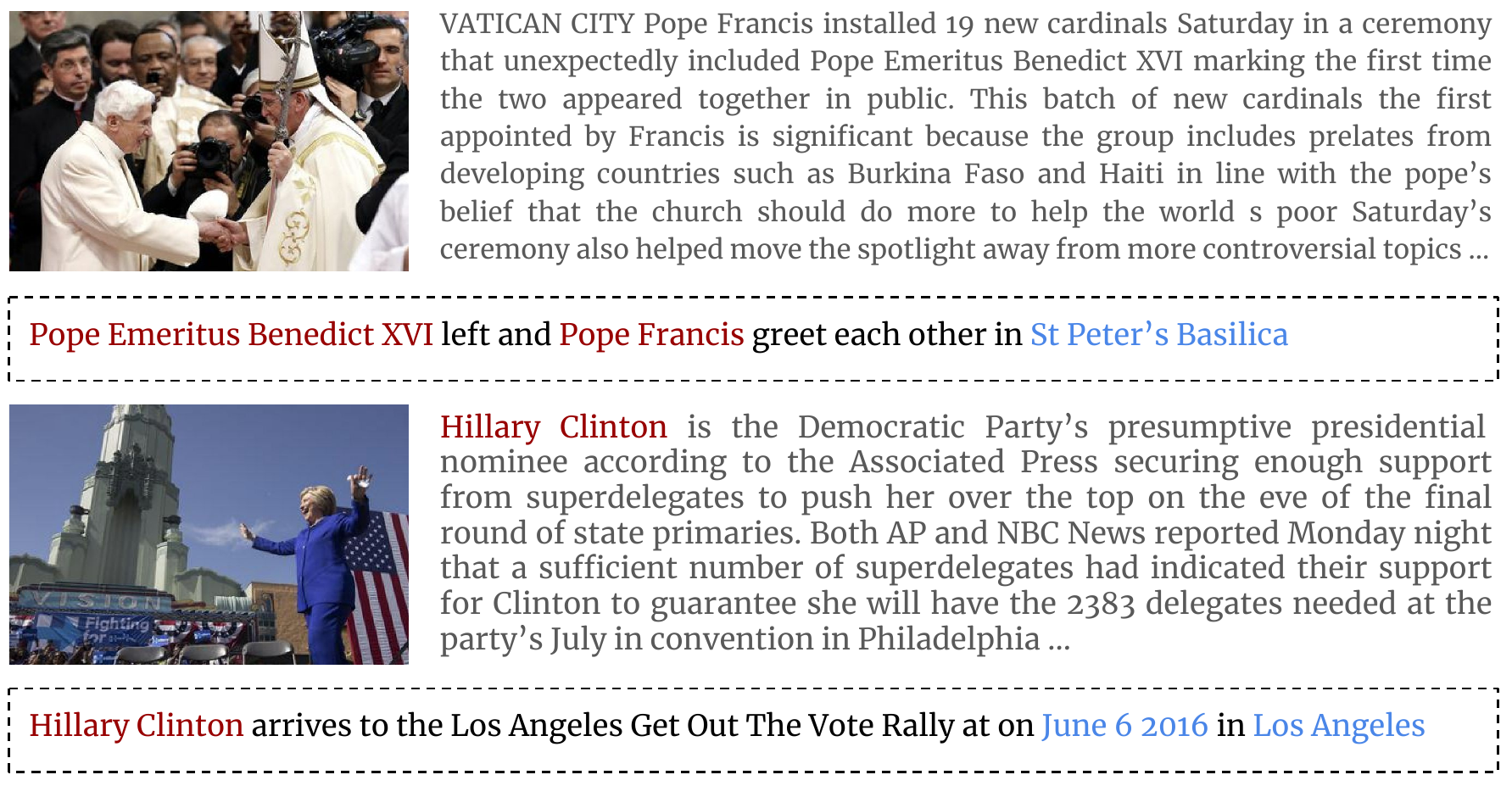}
    \caption{Examples of images from Visual News dataset and associated articles and captions.
    Named entities carrying important information are highlighted.
    %We highlight named entities which provide important information to readers.
    }
    \vspace{-0.2in}
    \label{fig:dataset_example}
\end{figure}

\begin{table*}[t]
\centering
\small
\begin{tabular}{c|c|c|ccccc}
\toprule
 &\multirow{2}{*}{\goodnews} & \multirow{2}{*}{NYTimes800k} & \multicolumn{5}{c}{\dataset~(ours)}\\
 & & & Guardian & BBC & USA & Wash. & Total\\
\midrule
Number of images & $462,642$ & $792,971$ & $602,572$ & $198,186$ & $151,090$ & $128,747$ & $1,080,595$ \\
Number of articles & $257,033$ & $444,914$ & $421,842$ & $97,429$ & $39,997$ & $64,096$ & $623,364$ \\
\midrule
% Average Article Length
Avg. Article Length & $451$ & $974$ & $787$ & $630$ & $700$ & $978$ & $773$ \\
Avg. Caption Length & $18$ & $18$ & $22.5$ & $14.2$ & $21.5$ & $17.1$ & $18.8$ \\
\midrule
%  Name Entities (Sentence)
\% of Sentences w/ NE  & $0.97$ & $0.96$ & $0.89$ & $0.85$ & $0.95$ & $0.92$ & $0.91$ \\
% Name Entities(Word)
\% of Words is NE & $0.27$ & $0.26$ & $0.18$ & $0.17$ & $0.22$ & $0.33$ & $0.22$ \\
Nouns & $0.16$ & $0.16$ & $0.20$ & $0.22$ & $0.17$ & $0.2$ & $0.19$ \\
Verbs & $0.09$ & $0.09$ & $0.10$ & $0.12$ & $0.08$ & $0.09$ & $0.09$ \\
Pronouns & $0.01$ & $0.01$ & $0.01$ & $0.01$ & $0.01$ & $0.01$ & $0.01$ \\
Proper nouns & $0.23$ & $0.22$ & $0.24$ & $0.18$ & $0.32$ & $0.28$ & $0.26$ \\
Adjectives & $0.04$ & $0.04$ & $0.06$ & $0.06$ & $0.05$ & $0.05$ & $0.06$ \\
\bottomrule
\end{tabular}
\caption{Statistics of news image datasets. \textit{"\% of Sentences w/ NE"} denotes the percentage of sentences containing named entities. \textit{"\% of Words is NE"} denotes the percentage of words that are used in named entities. }
\label{tab:dataset_summary}
\vspace{-0.05in}
\end{table*}

\begin{table*}[t]
    \centering
    \small
    \begin{subtable}[h]{0.45\textwidth}
        \begin{tabular}{ccccc}
\toprule
 & Guardian & BBC & USA & Wash. \\
\midrule
Guardian & $17745$ & $\bf 2345$ & $2048$ & $1997$ \\
BBC & $\bf2345$ & $12726$ & $1297$ & $1413$ \\
USA & $2048$ & $1297$ & $17013$ & $\bf2957$ \\
Wash. & $1997$ & $1413$ & $\bf2957$ & $16261$ \\
\bottomrule
       \end{tabular}
       \caption{PERSON entities.}
       \label{tab:PERSON}
    \end{subtable}
    \hspace{0.2cm}
    \begin{subtable}[h]{0.45\textwidth}
        \begin{tabular}{ccccc}
\toprule
 & Guardian & BBC & USA & Wash. \\
\midrule
Guardian & $2844$ & $814$ & $845$ & $\bf 910$ \\
BBC & $\bf814$ & $2038$ & $663$ & $731$ \\
USA & $845$ & $663$ & $3138$ & $\bf1162$ \\
Wash. & $910$ & $731$ & $\bf1162$ & $3221$ \\
\bottomrule
       \end{tabular}
       \caption{GPE entities.}
       \label{tab:GPE}
    \end{subtable}
    %\hfill
    \vspace{0.2cm}
    \\
    \begin{subtable}[h]{0.45\textwidth}
        \begin{tabular}{ccccc}
\toprule
 & Guardian & BBC & USA & Wash. \\
\midrule
Guardian & $8049$ & $\bf1146$ & $964$ & $958$ \\
BBC & $\bf1146$ & $6471$ & $701$ & $753$ \\
USA & $964$ & $701$ & $8487$ & $\bf1483$ \\
Wash. & $958$ & $753$ & $\bf1483$ & $8346$ \\
\bottomrule
       \end{tabular}
       \caption{ORG entities.}
       \label{tab:ORG}
    \end{subtable}
    \hspace{0.2cm}
    \begin{subtable}[h]{0.45\textwidth}
        \begin{tabular}{ccccc}
\toprule
 & Guardian & BBC & USA & Wash. \\
\midrule
Guardian & $3083$ & $\bf924$ & $776$ & $732$ \\
BBC & $\bf924$ & $2595$ & $682$ & $695$ \\
USA & $776$ & $682$ & $6491$ & $\bf1992$ \\
Wash. & $732$ & $695$ & $\bf1992$ & $3221$ \\
\bottomrule
       \end{tabular}
       \caption{DATE entities.}
       \label{tab:DATE}
    \end{subtable}
    
     \caption{Number of common named entities between different source agencies in Visual News dataset. "PERSON", "GPE", "ORG", and "DATE" are the top 4 most frequent named entity types. BBC has more common named entities with The Guardian than with USA Today and The Washington Post.}
     \label{tab:overlap}
\end{table*}

\label{sec:intro}

\section{Related Work}
Image captioning has gained increased attention, with remarkable results in recent benchmarks. A popular paradigm~\cite{vinyals2015show, karpathy2015deep, donahue2015long} uses a convolutional neural network as the image encoder and generates captions using a recurrent neural network (RNN) as the decoder. The seminal work of~\citet{xu2015show} proposed to attend to different image patches at different time steps and~\citet{lu2017knowing} improved this attention mechanism by adding an option to sometimes not to attend to any image regions. Other extensions include attending to semantic concept proposals~\cite{you2016image}, imposing local representations at the object level~\cite{li2017image} and a bottom-up and top-down attention mechanism to combine object and other salient image regions~\cite{anderson2018bottom}.

News image captioning is a challenging task because the captions often contain named entities. Prior work has attempted this task by drawing contextual information from the accompanying articles. \citet{tariq2016context} select the most representative sentence from the article; \citet{ramisa2017breakingnews} encode news articles using pre-trained word embeddings and concatenate them with CNN visual features to feed into an LSTM~\cite{HochSchm97}; \citet{lu2018entity} propose a template-based method in order to reduce the vocabulary size and then later retrieves named entities from auxiliary data; \citet{biten2019good} also adopt a template-based method but extract named entities by attending to sentences from the associated articles. \citet{zhao-etal-2019-informative} also tries to generate more informative image captions by integrating external knowledge. \citet{tran2020transform} proposes a transformer method to generates captions for images embedded in news articles in an end-to-end manner. 
In this work, we propose a novel Transformer based model to enable more efficient end-to-end news image captioning.

\begin{figure}[t]
    \centering
      \includegraphics[width=0.43\textwidth]{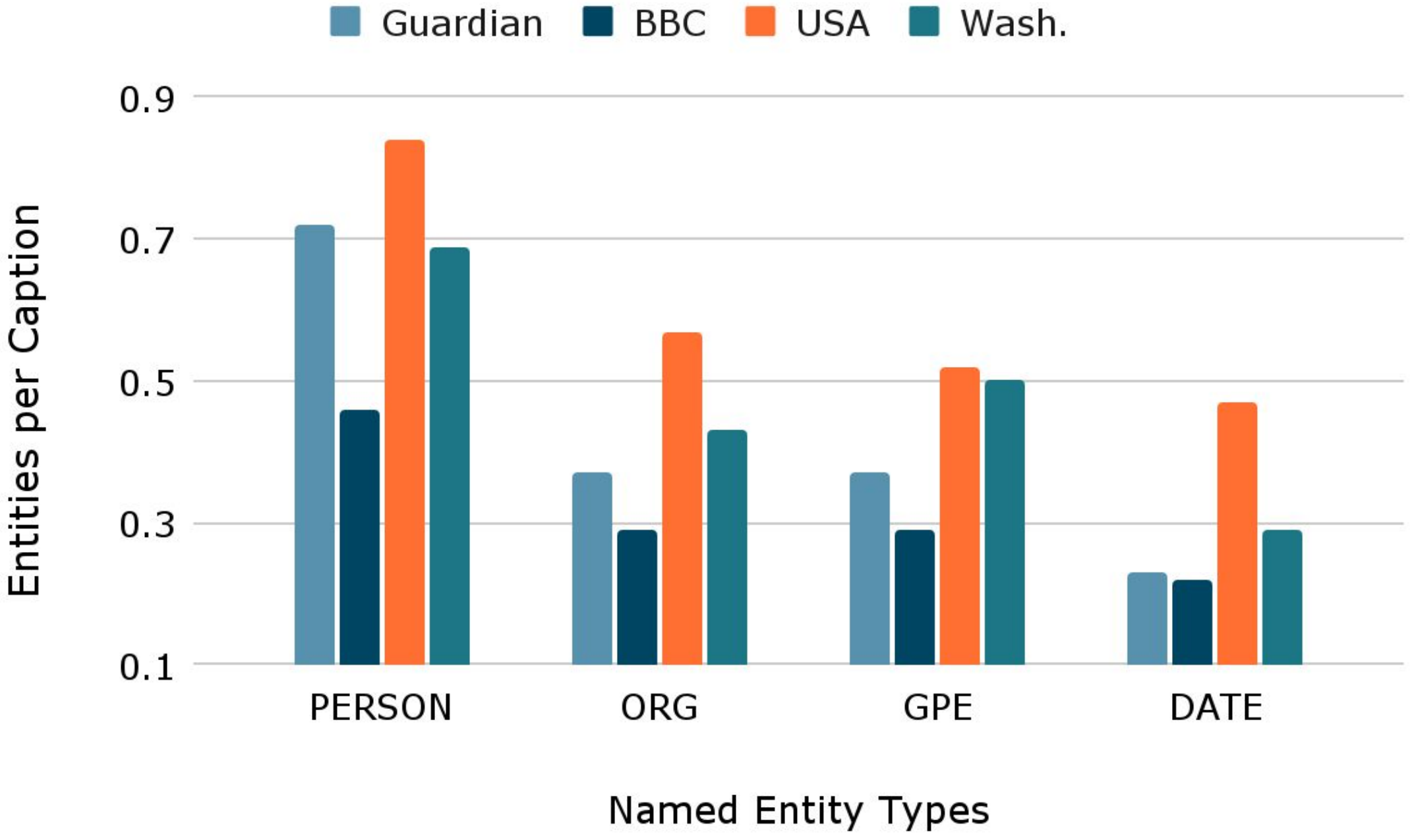}
     %\vspace{-3pt}
    \caption{
    Average count of named entities per caption. We select the top 4 most frequent named entity types in our Visual News dataset. For example, in The Guardian, there are on average $0.72$ PERSON entities per caption while it is $0.46$ for BBC. We see that each agency employs a distinct captioning style.
    }
    \vspace{-0.2in}
    \label{fig:diversity_analysis1}
\end{figure}

\begin{table}[t]
\centering
\small
\begin{tabular}{ccccc}
\toprule
 & Guardian & BBC & USA & Wash. \\
\midrule
Guardian & $1.0$ & $0.6$ & $0.6$ & $0.7$ \\
BBC & $1.9$ & $1.6$ & $1.7$ & $0.7$ \\
USA & $1.3$ & $1.2$ & $3.7$ & $2.7$ \\
Wash. & $1.2$ & $1.2$ & $2.0$ & $2.5$ \\
\bottomrule
\end{tabular}
\caption{CIDEr scores of the same captioning model on different train (row) and test (columns) splits. News images and captions from different agencies have different characters, leading to a performance decrease when training set and test set are not from the same agency. }
\label{tab:cross_analysis}
\end{table}

\label{sec:related_work}

\section{Our \dataset~Dataset}
\dataset~comprises news articles, images, captions, and other metadata from four news agencies: The Guardian, BBC, USA Today, and The Washington Post. To maintain quality, we first filter out images whose height or width is smaller than $180$ pixels. We then keep examples with a caption length between $5$ and $31$ words. Figure \ref{fig:dataset_example} shows some examples from \dataset. Although only images, captions, and articles are used in our experiments, \dataset~provides other metadata, such as article title, author, and geo-location.

\begin{figure*}[t!]
    \centering
      \includegraphics[width=0.9\textwidth]{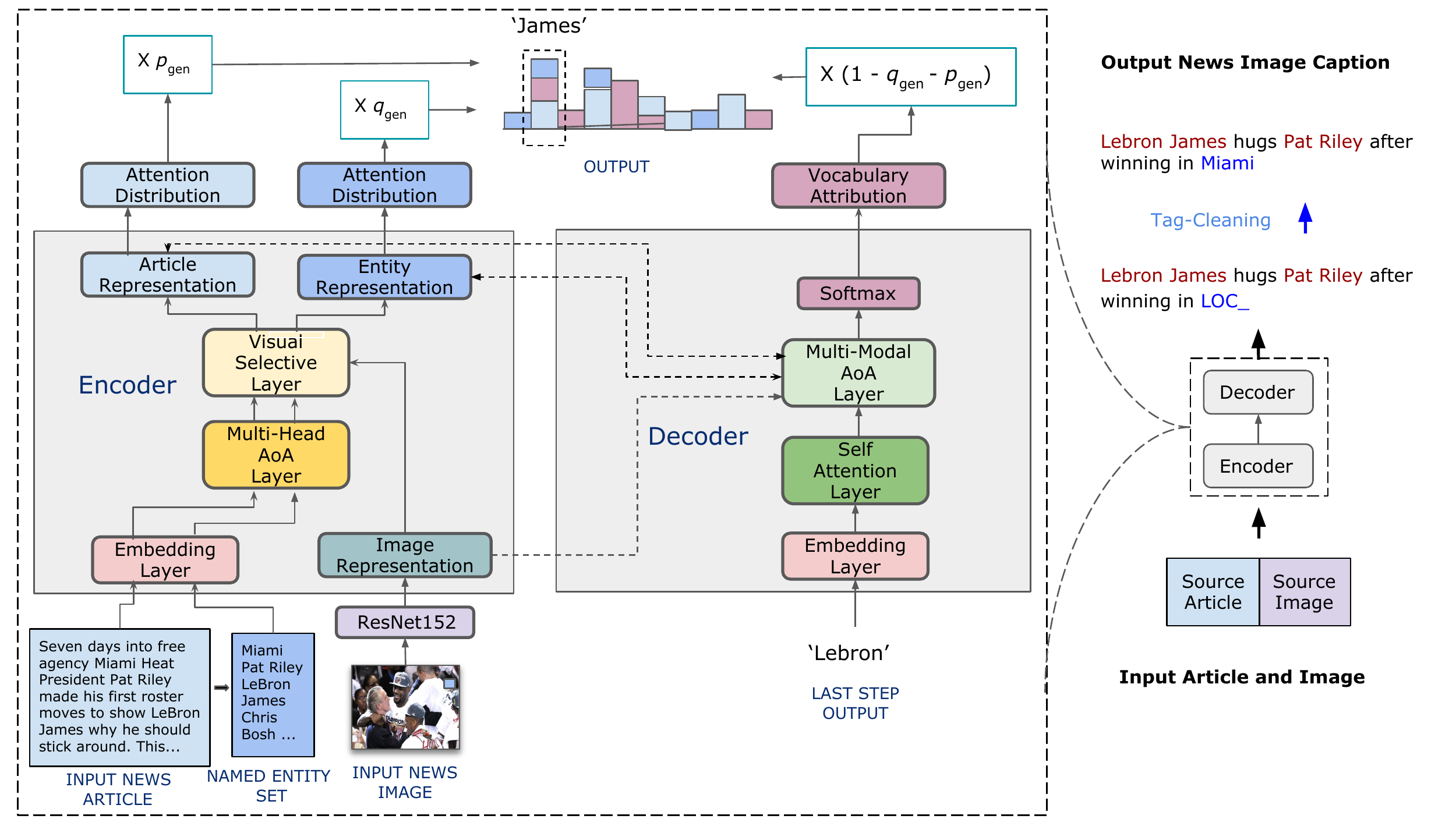}
    \caption{Overview of our model. Left: Details of the encoder and decoder; Right: The workflow of our model. The input news article and news image are fed into the encoder-decoder system. The blue arrow denotes the Tag-Cleaning step, which is a post-processing step to further improve the result during testing. Multi-Head AoA Layer means our Multi-Head Attention on Attention Layer. Multi-Modal AoA Layer means our Multi-Modal Attention on Attention Layer. Self Attention Layer denotes our Masked Multi-Head Attention on Attention Layer.}
    \label{fig:model}
\end{figure*}

We summarize the difference between \dataset~and other popular news image datasets in Table~\ref{tab:dataset_summary}. 
Compared to other recent news captioning datasets, such as~\goodnews~\cite{biten2019good} and NYTimes800k~\cite{tran2020transform}, \dataset~has two advantages. First, \dataset~has the largest number of images and articles. It contains over one million images and more than $600,000$ articles. Second, \dataset~is more diverse, since it contains articles from four news agencies. For example, the average caption length of BBC is only $14.2$ while for The Guardian it is $22.5$. In addition, only 18\% of the tokens in The Guardian are named entities while for The Washington Post it is 33\%. 

Figure~\ref{fig:diversity_analysis1} shows the average count of named entity types in captions from each agency. For instance, USA Today has on average $0.84$ "PERSON" entities per caption while BBC has only $0.46$. The Washington Post has $0.29$ "DATE" entities whereas USA Today has $0.47$. We also randomly select $50,000$ captions from each agency and calculate their unique named entities to see how many they have in common with each other (as summarized in Table~\ref{tab:overlap}). For example, BBC has more common named entities with The Guardian than USA Today and The Washington Post. USA Today shares more named entities of the same type with The Washington Post.

To further demonstrate the diversity in \dataset, we train a Show and Tell~\cite{vinyals2015show} captioning model on $100,000$ examples from a certain agency and test it on $10,000$ examples from the other agencies. We report CIDEr scores in Table~\ref{tab:cross_analysis}. A model trained on USA Today achieves a $3.7$ score on USA Today test set but only $0.6$ on The Guardian test set.\footnote{CIDEr scores are low since we directly use a baseline captioning method which is not designed for news images.} This gap also indicates that \dataset~is more diverse and also more challenging.

\label{sec:dataset}

\section{Method}
Figure~\ref{fig:model} presents an overview of \dataset~Captioner. We first introduce the image encoder and the text encoder.
% which consists of a Hybrid Embedding Layer, a Multi-Head Attention on Attention (AoA) \cite{huang2019attention} layer, and a Visual Selective Layer.
We then explain the decoder in section~\ref{section:4.3}.
% \ref{section:4.4} Multi-Head Pointer-Generator Module to calculate the final distribution.
To solve the out-of-vocabulary issue, we propose Tag-Cleaning in section~\ref{sec:post}.
% as a post processing step.
% to retrieve named entities from the article in testing.

\subsection{Image Encoder}
\label{section:4.1}
We use a ResNet152 \cite{he2016deep} pretrained on ImageNet~\cite{deng2009imagenet} to extract visual features. The output of the convolutional layer before the final pooling layer gives us a set of vectors corresponding to different patches in the image. Specifically, we obtain features $V=\{v_1, \dots, v_K\}, v_i\in\mathbb{R}^{D}$ from every image $I$, where $K=49$ and $D=2048$.
With these features, we can selectively attend to different regions at different time steps.

\subsection{Text Encoder} 
\label{section:4.2}
As the length of the associated article could be very long, we focus on the first $300$ tokens in each article following \cite{see2017get}. We also used the spaCy~\cite{honnibal2017spacy} named entity recognizer to extract named entities from news articles inspired by \citet{li2018guiding}.
We encode the first $300$ tokens and the extracted named entities using the same encoder. Given the input text $T=\{t_1, \dots, t_L\}$ where $t_i$ denotes the i-th token in the text and $L$ is the text length, we use the following layers to obtain textual features:

\vspace{2mm}
\noindent\textbf{Word Embedding and Position Embedding.}
% We encode the input tokens as well as the position information using a Hybrid Embedding Layer. 
For each token $t_i$, we first obtain a word embedding $w_i\in\mathbb{R}^{H}$ and positional embedding $p_i\in\mathbb{R}^{H}$ through two embedding layers, $H$ is the hidden state size and is set to $512$.  
% Since the Transformer is a nonrecurrent model, we also assign a
% We first obtain  from a lookup table \textcolor{red}{how you get $p_i$?} for the i-th element in a sequence. 
To better model the relative position relationships, we further feed position embeddings into an LSTM \cite{HochSchm97} to get the updated position embedding $p^l_i\in\mathbb{R}^{H}$.
% \textcolor{red}{I think you get the position embedding $p_i$ from a lookup table as BERT did. However, i am still confused about the difference between $p_i$ and $p^l_i$, I changed D into H since you have D in section 4.1}. 
We then add up $p^l_i$ and $w_i$ to obtain the final input embedding $w'_i$.
\begin{align}
p^l_i &= \text{LSTM}(p_i),\\
w'_i &= w_i + p^l_i.
\end{align}

\vspace{2mm}
\noindent\textbf{Multi-Head Attention on Attention Layer.} 
The Multi-Head Attention Layer \cite{vaswani2017attention} operates on three sets of vectors: queries $Q$, keys $K$ and values $V$, and takes a weighted sum of value vectors according to a similarity distribution between $Q$ and $K$. In our implementation, for each query $w'_i$, $K$ and $Q$ are all input embeddings $T'$.
In addition, we have the "Attention on Attention" (AoA) module \cite{huang2019attention} to assist the generation of attended information:
\begin{subequations}
\label{eq:aoa}
    \begin{align}
    v_{att} &= \text{MHAtt}(w'_i,T', T'),\\
    g_{att} &= \sigma(W_g[v_{att}; T']),\\
    v'_{att} &= W_a[v_{att};T'],\\
        % \text{MHAtt}_{AoA}(Q,K,V) = g_{att} \odot v'_{att}\tag{$5$}\\
        % \Tilde{w}_i = \text{MHAtt}_{AoA}(t^H_i,T^H,T^H)\tag{$6$}.
    \Tilde{w}_i &= g_{att} \odot v'_{att},
    \end{align}
\end{subequations}
where $\odot$ represents the element-wise multiplication operation and $\sigma$ is the sigmoid function. $W_g$ and $W_a$ are trainable parameters.
% $t^M_i$ is the output embedding of the Multi-Head AoA Layer in the encoder.

\vspace{2mm}
\noindent\textbf{Visual Selective Layer.} One limitation of previous works \cite{tran2020transform, biten2019good} is that they separately encode the image and article, ignoring the connection between them during encoding. In order to generate representations that can capture contextual information from both images and articles, we propose a novel Visual Selective Layer which updates textual embeddings with a visual information gate: 
\begin{align}
\overline{T} &= \text{AvgPool}(\Tilde{T}),\\
g_v &= \text{tanh}(W_{v}(\text{MHAtt}_{\text{AoA}}(\overline{T},V,V)),\\
w^{*}_i &= g_{v} \odot \Tilde{w}_i,\\
w^a_i &= \text{LayerNorm}(w^{*}_i + \text{FFN}(w^{*}_i)),
\end{align}
where $\text{MHAtt}_\text{AoA}$ corresponds to Eq~\ref{eq:aoa}. To obtain fixed-length article representations, we apply the average pooling operation to get $\overline{T}$, which can be used as the query to attend to different regions of the image.
% Then we use $g_v$ to filter\textcolor{red}{what do you mean? weighted?} the source text representation. 
\text{FFN} is a two-layer feed-forward network with ReLU as the activation function. $w^a_i$ is the final output embedding from the text encoder. For the sake of simplicity, in the following text, we use $A$ = \{$a_1, \dots, a_L\}, a_i\in \mathbb{R}^{H}$ to represent the final embeddings ($w^a_i$) of article tokens, where $H$ is the embedding size and $L$ is the article length. Similarly,  $E$ = \{$e_1, \dots, e_M\}, e_i\in \mathbb{R}^{H}$ represent the final embeddings of extracted named entities, where $M$ is the number of named entities.

\subsection{Decoder}
\label{section:4.3}
Our decoder generates the next token conditioned on previously generated tokens and contextual information. 
We propose Masked Multi-Head Attention on Attention Layer to flexibly attend to the previous tokens and Multi-Modal Attention on Attention Layer to fuse contextual information.
We first use the encoder to obtain embeddings of ground truth captions $X$ = \{$x_0, \dots, x_N$\}, $x_i\in \mathbb{R}^{H}$, where $N$ is the caption length and $H$ is the embedding size.
Instead of using the Masked Multi-Head Attention Layer used in \citet{tran2020transform} to collect the information from past tokens, we use the more efficient Masked Multi-Head Attention on Attention Layer. At time step $t$, output embedding $x^a_t$ is used as the query to attend over context information:
\begin{align}
    x^a_t = \text{MHAtt}^\text{Masked}_\text{AoA}(x_t,X,X).
\end{align}

\vspace{2mm}
\noindent\textbf{Multi-Modal Attention on Attention Layer.} Our Multi-Modal AoA Layer contains three context sources: images ${V}$, articles $A$ and name entity sets $E$. We use a linear layer to resize features in $V$ into $\Tilde{V}$, where $\Tilde{v}\in \mathbb{R}^{512}$. In each step, $x^a_i$ is the query that attends over them separately:
\begin{align}
    V'_t &= \text{MHAtt}_\text{AoA}(x^a_t,\Tilde{V}, \Tilde{V}),\\
    A'_t &= \text{MHAtt}_\text{AoA}(x^a_t,A,A),\\
    E'_t &= \text{MHAtt}_\text{AoA}(x^a_t,E,E).
\end{align}

We combine the attended image feature $V'_t$, the attended article feature  $A'_t$ and the attended named entity feature $E'_t$, and feed them into a residual connection, layer normalization and a two-layer feed-forward layer FFN.
\begin{align}
    C_t &= V'_t + A'_t + E'_t,\\
    x'_t &= \text{LayerNorm}(x^a_t + C_t),\\
    x^*_t &= \text{LayerNorm}(x'_t + \text{FFN}(x'_t)),\\
    P_{s_t} &= \text{softmax}(x^*_t).
\end{align}
The final output $P_{s_t}$ will be used to predict token $s_t$ in the Multi-Head Pointer-Generator Module.

\vspace{2mm}
\noindent\textbf{Multi-Head Pointer-Generator Module.} For the purpose of obtaining more related named entities from the associated article and the extracted named entity set, we adapt the pointer-generator \cite{see2017get}.
Our pointer-generator contains two sources: the article and the named entity set.
We first generate $a^V$ and $a^E$ over the source article tokens and extracted named entities by averaging the attention distributions from the multiple heads of the Multi-Modal Attention on Attention layer in the last decoder layer. Next, $p_{gen}$ and $q_{gen}$ are calculated as two soft switches to choose between generating a word from the vocabulary distribution $P_{s_t}$, or copying words from the attention distribution $a^V$ or $a^E$:
\begin{align}
    p_{gen} &= \sigma(W_p([x_t;A'_t;V'_t])),\\
    q_{gen} &= \sigma(W_q([x_t;E'_t;V'_t])),
\end{align}
where $A'_i$, $V'_i$ and $E'_i$ are attended context vectors, $W_p$ and $W_q$ are learnable parameters, and $\sigma$ is the sigmoid function. $P^*_{s_i}$ provides us with the final distribution to predict the next word.
\begin{equation}
    \begin{split}
    P^*_{s_t} = p_{gen}a^V + q_{gen}a^E + \\
    (1 - p_{gen} - q_{gen})P_{s_t}.
    \end{split}
\end{equation}

Finally, our loss can be computed as the sum of the negative log-likelihood of the target word at each time step:
\begin{align}
Loss = -\sum_{t=1}^{N}\log P^*_{s_i}.
\end{align}

\subsection{Tag-Cleaning}
\label{sec:post}
To solve the out-of-vocabulary (\textit{OOV}) problem, we replace \textit{OOV} named entities with named entity tags instead of using a single ``UNK'' token, e.g. if ``John Paul Jones Arena'' is a \textit{OOV} named entity, we replace it with ``LOC\_'', which represents location entities. During testing, if the model predicts entity tags, we further replace those tags with specific named entities. More specifically, we select a named entity with the same entity category and the highest frequency from the named entity set.

\label{sec:model}

\section{Experiments}
\label{sec:experiment}
% This section describes the settings for network learning, evaluation metrics, baselines and competing methods, followed by results and discussions.
In this section, we first introduce details of implementation. Then baselines and competing methods will be discussed. Lastly, we present comprehensive experiment results on both the \goodnews~dataset and our \dataset~dataset. 
\subsection{Implementation Details}
% Our model is developed with PyTorch~\cite{paszke2017automatic}. 

\begin{table*}[t]
\setlength\tabcolsep{3pt}
\centering
\small
\begin{tabular}{lccccccccc}
\toprule
Model & Solve OOV & BLEU-4 & METEOR & ROUGE & CIDEr & P & R \\
\midrule
TextRank \cite{barrios2016variations}& \xmark &$2.1$ & $8.0$ & $12.0$ & $8.4$ & $4.1$ & $6.1$ \\
Show Attend Tell \cite{xu2015show}& \xmark
& $1.5$ & $4.6$ & $12.6$ & $11.3$ & $-$ & $-$ \\
Tough-to-beat \cite{biten2019good}& \xmark %Tag-Cleaning %\cite{biten2019good} 
& $1.7$ & $4.6$ & $13.2$ & $12.4$ & $4.9$ & $4.8$ \\
Pooled Embeddings \cite{biten2019good}& \xmark
& $2.1$ & $5.2$ & $13.5$ & $13.2$ & $5.3$ & $5.3$ \\
\midrule
Our Transformer & \xmark & $4.9$ & $7.7$ & $16.8$ & $45.6$ & $18.5$ & $16.1$ \\
Our Transformer+EG & \xmark & $5.0$ & $7.9$ & $17.4$ & $46.8$ & $19.2$ & $16.7$ \\
Our Transformer+EG+Pointer & \xmark & $5.1$ & $8.0$ & $17.7$ & $48.0$ & $19.3$ & $17.0$ \\
Our Transformer+EG+Pointer+VS & \xmark & $5.1$ & $8.1$ & $17.8$ & $48.6$ & $19.4$ & $17.1$ \\
Our Transformer+EG+Pointer+VS+TC & Tag-Cleaning & $\bf5.3$ & $\bf8.2$ & $\bf17.9$ & $\bf50.5$ & $\bf19.7$ & $\bf17.6$ \\
\bottomrule
\end{tabular}
\caption{News image captioning results (\%) on our \dataset~dataset. EG means adding the named entity set as another text source guiding the generation of captions. Pointer means pointer-generator module. VS means the Visual Selective Layer. TC means the Tag-Cleaning step.
%Across various metrics, our method outperforms existing methods or reaches a comparably good performance. Similarly, we observe a significant performance gain by adopting our proposed modules. EG means adding the named entity set as another text source guiding the generation of captions. Pointer means pointer-generator module. VS means the Visual Selective Layer. PE means adding our Position Embedding. TC means the Tag-Cleaning step.
}
\label{tab:VisualNews_result}
\end{table*}

\begin{table*}[t]
\setlength\tabcolsep{3pt}
\centering
\small
\begin{tabular}{clccccccccc}
\toprule
\parbox[t]{4mm}{\multirow{8}{*}{\rotatebox[origin=c]{90}{GoodNews}}}
 & Model & Solve OOV & BLEU-4 & METEOR & ROUGE & CIDEr & P & R \\
\midrule
& TextRank \cite{barrios2016variations} & \xmark &$1.7$ & $7.5$ & $11.6$ & $9.5$ & $1.7$ & $5.1$ \\
& Show Attend Tell \cite{xu2015show}& \xmark& $0.7$ & $4.1$ & $11.9$ & $12.2$ & $-$ & $-$ \\
& Tough-to-beat \cite{biten2019good} & \xmark & $0.8$ & $4.2$ & $11.8$ & $12.8$ & $9.1$ & $7.8$ \\
& Pooled Embeddings \cite{biten2019good} & \xmark & $0.8$ & $4.3$ & $12.1$ & $12.7$ & $8.2$ & $7.2$ \\
& Transform and Tell \cite{tran2020transform}& BPE & $6.0$ & $-$ & $21.4$ & $53.8$ & $22.2$ & $18.7$ \\\parbox[t]{3mm}{\multirow{8}{*}{\rotatebox[origin=c]{90}{NYTimes800k}}}
& \bf{Visual News Captioner} & Tag-Cleaning & $\bf6.1$ & $\bf8.3$ & $\bf21.6$ & $\bf55.4$ & $\bf22.9$ & $\bf19.3$\\
\midrule
% \midrule
% \parbox[t]{4mm}{\multirow{8}{*}{\rotatebox[origin=c]{90}{NYTimes800k}}
& TextRank \cite{barrios2016variations} & \xmark &$1.9$ & $7.3$ & $11.4$ & $9.8$ & $3.6$ & $4.9$ \\
& Tough-to-beat \cite{biten2019good} & \xmark & $0.7$ & $4.2$ & $11.5$ & $12.5$ & $8.9$ & $7.7$ \\
& Pooled Embeddings \cite{biten2019good} & \xmark & $0.8$ & $4.1$ & $11.3$ & $12.2$ & $8.6$ & $7.3$ \\
& Transform and Tell \cite{tran2020transform}& BPE & $6.3$ & $-$ & $21.7$ & $54.4$ & $24.6$ & $22.2$ \\
& \bf{Visual News Captioner} & Tag-Cleaning & $\bf6.4$ & $\bf8.1$ & $\bf21.9$ & $\bf56.1$ & $\bf24.8$ & $\bf22.3$ \\
\bottomrule
\end{tabular}
\caption{News image captioning results (\%) on \goodnews~and NYTimes800k dataset.
%Across various metrics, our method outperforms existing methods or reaches a comparably good performance. EG means adding the named entity set as another text source guiding the generation of captions. Pointer means pointer-generator module. VS means the Visual Selective Layer. PE means adding our Position Embedding. TC means the Tag-Cleaning step.
}
\label{tab:GoodNews_result}
\end{table*}

\begin{table}[t]
\setlength\tabcolsep{3pt}
\centering
\small
\begin{tabular}{lcc}
\toprule
Model & Number of Parameters\\
\midrule
Transform and Tell \cite{tran2020transform} & 200M\\
\dataset~Captioner & \textbf{93M}\\
Visual News Captioner (w/o Pointer) & 91M\\
Visual News Captioner (w/o EG) & 91M\\
\bottomrule
\end{tabular}
\caption{
% Number of training parameters.
We compare the number of training parameters of our model variants with Transform and Tell \citep{tran2020transform}. Note that our proposed \dataset Captioner is much more lightweight.
}
\label{tab:post_ablation}
\end{table}

\vspace{1mm}
\noindent\textbf{Datasets.}~We conduct experiments on three large-scale news image datasets:~\goodnews,~\NYT~and~\dataset. For \goodnews~and \NYT, we follow the setting from the original paper. For \dataset, we randomly sample $100,000$ images from each news agency, leading to a training set of $400,000$ samples. 
Similarly, we get a $40,000$ validation set and a $40,000$ test set, both evenly sampled from four news agencies. 

Throughout our experiments, we first resize images to a $256 \times 256$ resolution, and randomly crop patches to a size of $224 \times 224$ as input. To preprocess captions and articles, we remove noisy HTML labels, brackets, non-ASCII characters, and some special tokens. We use spaCy’s named entity recognizer \cite{honnibal2017spacy} to recognize named entities in both captions and articles.

\vspace{1mm}
\noindent\textbf{Model Training.} We set the embedding size $H$ to 512. For dropout layers, we set the dropout rate as $0.1$. Models are optimized using Adam \cite{kingma2014adam} with warming up learning rate set to $0.0005$. We use a batch size of $64$ and stop training when the CIDEr~\cite{vedantam2015cider} score on the dev set is not improving for $20$ epochs. Since we replace $OOV$ named entities with tags, we add 18 named entity tags provided by spaCy into our vocabulary, including "PERSON\_", "LOC\_", "ORG\_", "EVENT\_", etc.

% \vspace{1mm}
\noindent\textbf{Evaluation Metrics.} 
Following previous literature, we evaluate model performance on two categories of metrics.
% We conduct two main evaluations of our model. 
To measure the overall similarity between generated captions and ground truth, we report BLEU-4 \cite{papineni2002bleu}, METEOR \cite{denkowski2014meteor}, ROUGE \cite{ganesan2018rouge} and CIDEr \cite{vedantam2015cider} scores. Among these scores, CIDEr is the most suitable for measuring performance in news captioning since it downweighs stop words and focuses more on uncommon words through a TF-IDF weighting mechanism. On the other hand, we compute the precision and recall scores for named entities to evaluate the model ability to predict named entities.
\begin{figure*}[t]
    \centering
      \includegraphics[width=\textwidth]{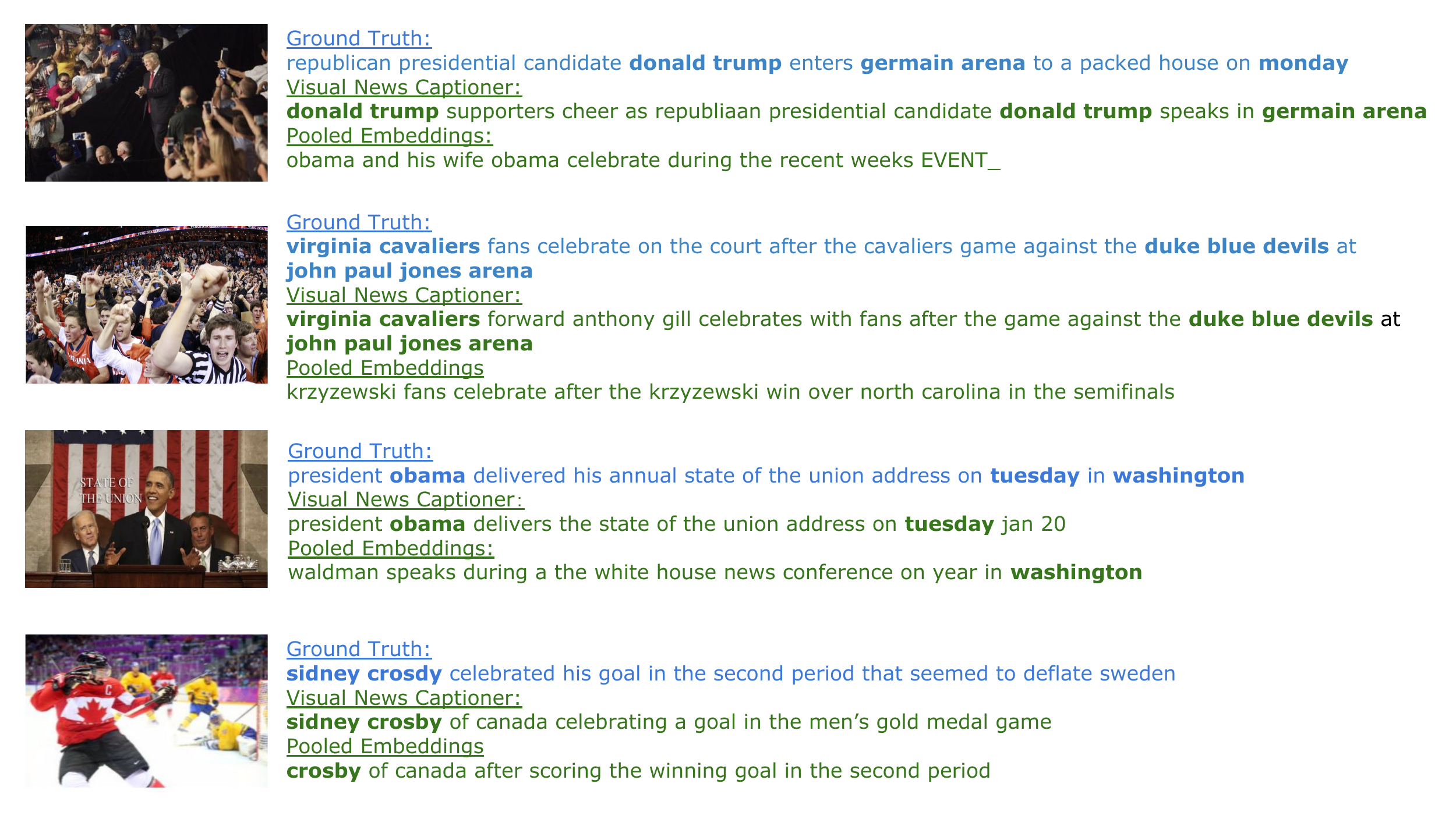}
    \caption{
    Examples of captions generated by different models. The first three are from \dataset~and the last one is from \goodnews. Correct named entities are highlighted in bold. Our \dataset~Captioner is able to predict the named entities more accurately and completely than the competing method.
    %We show captions predicted by our method along with outputs from our baselines as well as ground truth captions. Notice our \dataset-Captioner is able to predict the named entities more accurately and completely while Pool Embeddings \cite{biten2019good} generates only some of the named entities. 
    }
    \label{fig:qualitative_example}
\end{figure*}
% \vspace{-2mm}
\subsection{Competing Methods and Model Variants}
We compare our proposed Visual News Captioner with various baselines and competing methods.

\vspace{1mm}
\noindent\textbf{TextRank}~\cite{barrios2016variations} is a graph-based extractive summarization algorithm. This baseline only takes the associated articles as input. 

\vspace{1mm}
\noindent\textbf{Show Attend Tell} \cite{xu2015show} tries to attend to certain image patches during caption generation. This baseline only takes images as input.  

\vspace{1mm}
\noindent\textbf{Pooled Embeddings} and \textbf{Tough-to-beat}~\citep{arora2016simple} are two template-based models proposed in~\citet{biten2019good}.
%\footnote{Named as Avg+CtxIns and TBB+AttIns in the original paper.}.
They try to encode articles at the sentence level and attend to certain sentences at different time steps.
\textit{Pooled Embeddings} computes sentence representations by averaging word embeddings and adopts context insertion in the second stage.
\textit{Tough-to-beat} obtains sentence representations from the tough-to-beat method introduced in \citet{arora2016simple} and uses sentence level attention weights~\cite{biten2019good} to insert named entities.

\vspace{1mm}
\noindent\textbf{Transform and Tell}~\cite{tran2020transform} is the transformer-based attention model, which uses a pretrained RoBERTa \cite{liu2019roberta} model as the article encoder and a transformer as the decoder. It uses byte-pair encoding (BPE) to represent out-of-vocabulary named entities. 

\noindent\textbf{Visual News Captioner} is our proposed model, which is
based on transformer~\cite{vaswani2017attention}. Our transformer adopts Multi-Head Attention on Attention (AoA).
% \textit{Transformer*}~basically follows the transformer in \cite{vaswani2017attention} except we replace the Multi-Head Attention with our Multi-Head AoA;
EG (Entity-Guide) adds named entities as another text source to help predict named entities more accurately.
VS (Visual Selective Layer) tries to strengthen the connection between the image and text.
Pointer stands for the updated multi-head pointer-generator module.
% All of the above models work directly on raw captions without dealing with \textit{OOV} problem. 
To overcome the limitation of a fixed-size vocabulary, we examine TC, the Tag-Cleaning operation handling the \textit{OOV} problem. 

\subsection{Results and Discussion}
Table~\ref{tab:GoodNews_result} and Table~\ref{tab:VisualNews_result} summarize our quantitative results on the \goodnews~, \NYT~ and \dataset~datasets respectively. 
%More specifically, we have following findings from our experimental results: 
% On both datasets, our \dataset-Captioner model outperforms the state-of-the-art methods or reaches a comparably good performance across multiple metrics. 
On \goodnews~and \NYT, our Visual News Captioner outperforms the state-of-the-art methods on all 6 metrics. On our \dataset~dataset, our model outperforms baseline methods by a large margin, from $13.2$ to $50.5$ in CIDEr score.
In addition, as revealed by Table~\ref{tab:post_ablation}, our final model  outperforms \textit{Transform and Tell (transformer)} with much fewer parameters. This demonstrates that our proposed model is able to generate better captions in a more efficient way.

Our Entity-Guide (EG) brings improvement in all datasets, demonstrating that the named entity set contains key information guiding the generation of news captions. In addition, our pointer-generator mechanism builds a stronger connection between the final distribution of the predicted tokens and the Multi-Modal AoA Layer. More importantly, our Visual Selective Layer (VS) improves the caption generation results by providing extra visual context to text features.

Furthermore, our Tag-Cleaning (TC) method is able to effectively retrieve uncommon named entities and thus improves the CIDEr score by $1.3\%$ on the \dataset~datasets. We present qualitative results of different models on both datasets in Figure~\ref{fig:qualitative_example}. Our model shows the ability to generate more accurate named entities.

% We also observe that both our models and \textit{Transform and Tell} methods are directly trained on raw captions but generate better captions than all template-based methods.
We also observe that our models and \textit{Transform and Tell} methods achieve the best performances are directly trained on raw captions rather than following a two-stage template-based manner.
Although template-based methods normally handle a much smaller vocabulary, these methods also suffer from losing rich contextual information brought by uncommon named entities. 

The performance on the \goodnews~dataset and \NYT~dataset is better compared to the results on \dataset. This is because our \dataset~dataset is more challenging in terms of diversity. Our \dataset~ dataset is collected from multiple news agencies, thus, covers more topics and has more diverse language styles.

\section{Conclusion and Future Work}
In this paper, we study the task of news image captioning. First, we construct \dataset, the largest news image captioning dataset consisting of over one million images with accompanying articles, captions, and other metadata. Furthermore, we propose Visual News Captioner, an entity-aware captioning method leveraging both visual and textual information. We validate the effectiveness of our method on three datasets through extensive experiments. Visual News Captioner outperforms state-of-the-art methods across multiple metrics with fewer parameters.
Moreover, our \dataset~dataset can potentially be adapted to other NLP tasks, such as abstractive text summarization and fake news detection. We hope this work paves the way for future studies in news image captioning as well as other related research areas.
\label{sec:conclusion}

\section*{Acknowledgements}
This work was supported in part by an NVIDIA hardware Grant. We are also thankful for the feedback from anonymous reviewers of this paper.
\label{sec:Acknowledgements}

\bibliography{anthology,custom}
\bibliographystyle{acl_natbib}

% \appendix
% \clearpage
% \section{Appendix}
% \input{sections/appendix.tex}
% \label{sec:appendix}
% This is an appendix.

\end{document}